%% file: root.tex
\documentclass[letterpaper, 10 pt, conference]{ieeeconf} % Comment this line out if you need a4paper

\IEEEoverridecommandlockouts                              % This command is only needed if 
\usepackage{graphics} % for pdf, bitmapped graphics files
\usepackage{epsfig} % for postscript graphics files
\usepackage{times} % assumes new font selection scheme installed
\usepackage{amsmath} % assumes amsmath package installed
\usepackage{amssymb}  % assumes amsmath package installed
\usepackage{biblatex}
\usepackage{multirow}
\usepackage{multicol}
\usepackage{hyperref}
\usepackage{booktabs}
\usepackage{enumerate}
\usepackage{color}
\usepackage{tabularx}
\usepackage{wrapfig}
\usepackage{hyperref}
\usepackage{pifont}
\usepackage{algpseudocode}
\usepackage{algorithm}
\usepackage{amsmath}
\usepackage{adjustbox}
\usepackage[usenames,dvipsnames,table]{xcolor}
\hypersetup{
    colorlinks=true,
    linkcolor=MidnightBlue,
    filecolor=magenta,      
    urlcolor=MidnightBlue,
    citecolor=MidnightBlue,
} 
\usepackage[font=small,labelfont=bf]{caption}
\newcommand{\todo}[1]{\textcolor{red}{TODO #1}}

\definecolor{darkgrey}{rgb}{0.5,0.5,0.5}

\newcommand\red[1]{{{#1}}}

\title{
\LARGE\bf FoAR: Force-Aware Reactive Policy for\\ Contact-Rich Robotic Manipulation
}

\author{Zihao He$^{*}$, Hongjie Fang$^{*}$,  Jingjing Chen, Hao-Shu Fang$^{\dagger}$ and Cewu Lu$^{\dagger}$ \\ %
Shanghai Jiao Tong University % <-this % stops a space
\thanks{$^{*}$ Equal Contribution.}
\thanks{$^\dagger$ Hao-Shu Fang and Cewu Lu are the corresponding authors.} %
\thanks{Emails: \{he0610, galaxies, jjchen20\}@{sjtu.edu.cn}, fhaoshu@gmail.com, lucewu@sjtu.edu.cn}%
}

\begin{document}

\maketitle
\thispagestyle{empty}
\pagestyle{empty}

%%%%%%%%%%%%%%%%%%%%%%%%%%%%%%%%%%%%%%%%%%%%%%%%%%%%%%%%%%%%%%%%%%%%%%%%%%%%%%%%
\begin{abstract}
Contact-rich tasks present significant challenges for robotic manipulation policies due to the complex dynamics of contact and the need for precise control. Vision-based policies often struggle with the skill required for such tasks, as they typically lack critical contact feedback modalities like force/torque information. To address this issue, we propose FoAR, a force-aware reactive policy that combines high-frequency force/torque sensing with visual inputs to enhance the performance in contact-rich manipulation. Built upon the RISE policy, FoAR incorporates a multimodal feature fusion mechanism guided by a future contact predictor, enabling dynamic adjustment of force/torque data usage between non-contact and contact phases. Its reactive control strategy also allows FoAR to accomplish contact-rich tasks accurately through simple position control. Experimental results demonstrate that FoAR significantly outperforms all baselines across various challenging contact-rich tasks while maintaining robust performance under unexpected dynamic disturbances. Project website: \href{https://tonyfang.net/FoAR/}{https://tonyfang.net/FoAR/}.
\end{abstract}

%\begin{IEEEkeywords}
%Force/Torque,  Contact-Rich Manipulation, Imitation Learning
%\end{IEEEkeywords}

%%%%%%%%%%%%%%%%%%%%%%%%%%%%%%%%%%%%%%%%%%%%%%%%%%%%%%%%%%%%%%%%%%%%%%%%%%%%%%%%
\input{1_introduction}
\input{2_related_works}

\input{3_method}
\input{4_experiments}

\input{6_conclusions}


%%%%%%%%%%%%%%%%%%%%%%%%%%%%%%%%%%%%%%%%%%%%%%%%%%%%%%%%%%%%%%%%%%%%%%%%%%%%%%%%

%%%%%%%%%%%%%%%%%%%%%%%%%%%%%%%%%%%%%%%%%%%%%%%%%%%%%%%%%%%%%%%%%%%%%%%%%%%%%%%%

%%%%%%%%%%%%%%%%%%%%%%%%%%%%%%%%%%%%%%%%%%%%%%%%%%%%%%%%%%%%%%%%%%%%%%%%%%%%%%%%

\input{acknowledgement}

\printbibliography

% not for RAL, but can be included in arXiv version.
\input{appendix}

\end{document}

%% file: 1_introduction.tex
\section{Introduction}\label{sec:introduction}

Contact-rich manipulation is an essential field in robotics, involving tasks that require sustained, intricate contact with objects or environments~\cite{suomalainen2022survey}. Such tasks, including assembly~\cite{furniturebench, mimictouch}, wiping~\cite{acp, maniwav}, and peeling~\cite{chen2024vegetable, liu2024force}, are inherently challenging due to the complex dynamics of force and precise control required. Unlike simple pick-and-place operations~\cite{transporter}, contact-rich manipulation demands nuanced interaction and real-time adaptation to variations in object properties. As a result, developing effective algorithms and learning models for contact-rich manipulation is crucial for advancing robotic dexterity, enabling more versatile, autonomous, and interactive robot systems.

\begin{figure}[t]
    \centering
    \includegraphics[width=\linewidth]{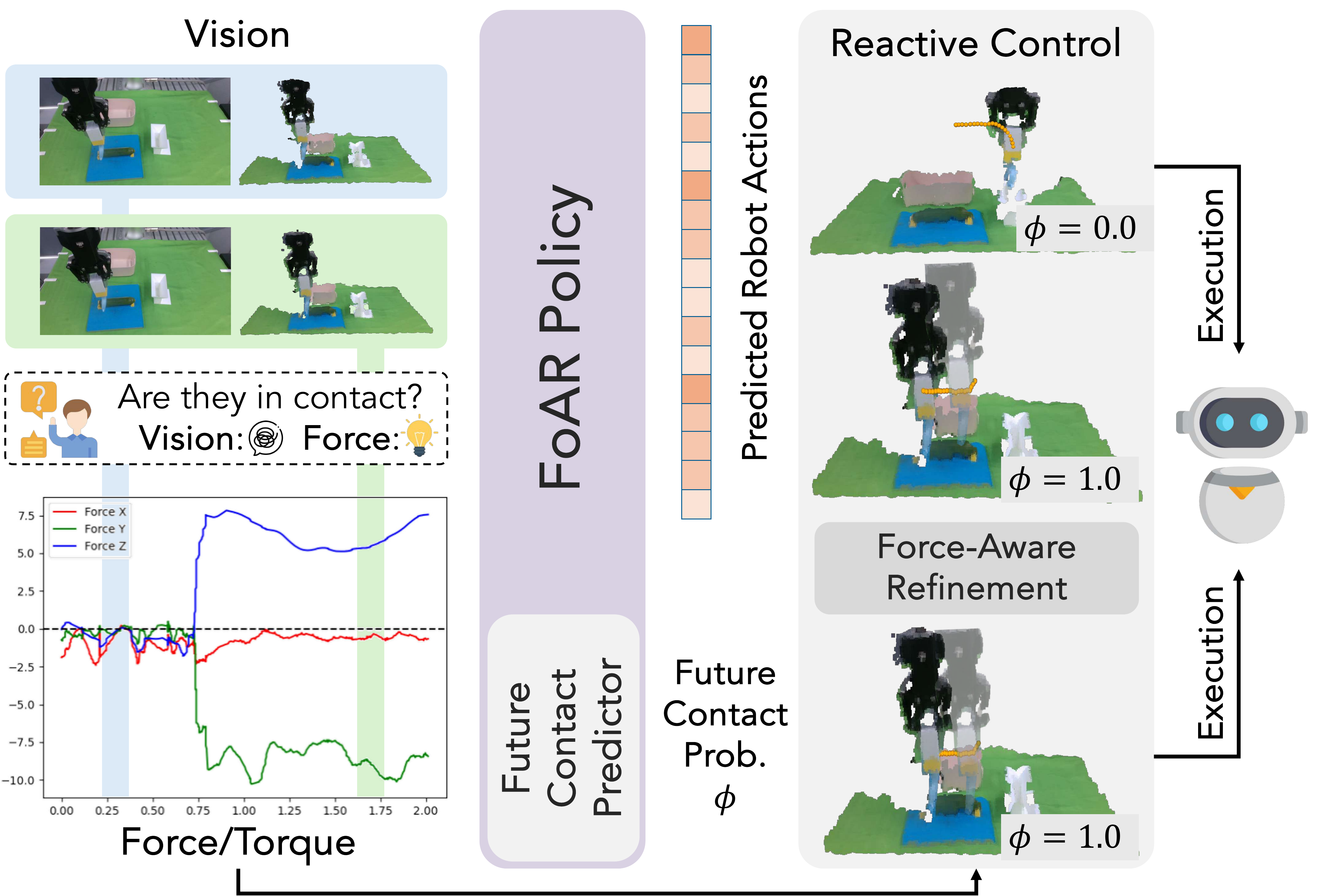}
    \caption{\textbf{Overview of the FoAR Policy for Contact-Rich Robotic Manipulations.} Vision alone struggles to distinguish contact from non-contact states in contact-rich tasks, underscoring the need for integrating force/torque information. Our FoAR policy combines vision and force/torque inputs to predict robot actions along with a future contact probability $\phi$. Reactive control then refines actions dynamically based on current and predicted future contact states, enabling precise, force-aware manipulations for contact-rich tasks.}
    \label{fig:teaser}\vspace{-0.3cm}
\end{figure}

In recent years, significant progress has been made in vision-based robotic manipulation policies~\cite{rt1, diffusionpolicy, oxe, openvla, octo, rise, cage, act, rt2}. However, these policies often fall short of achieving the dexterity required for contact-rich manipulations, as they typically lack crucial contact feedback, such as force/torque and tactile information. This limitation hinders the robot's ability to perceive contact states and understand physical interactions, thus constraining its manipulation capabilities, as illustrated in Fig.~\ref{fig:teaser} (left).

To address the limitations of pure vision-based policies, recent approaches have incorporated additional modalities such as audio~\cite{playtothescore, seehearfeel, maniwav, hearingtouch}, tactile~\cite{playtothescore, huang20243d, seehearfeel, eyesight_hand}, and force/torque~\cite{vtt, makingsensevisiontouch, tacdiffusion, zhou2024admittance} into the policy framework. These multi-modal policies offer promising avenues for advancing robotic manipulation by providing richer feedback about interactions, enabling robots to handle contact-rich tasks with greater precision and adaptability.

Nevertheless, audio and similar indirect sensing modalities are typically vulnerable to background noise~\cite{maniwav}, complicating signal processing and reducing reliability in real-world applications. Additionally, they often fail to deliver detailed information about contact dynamics between robots and objects, constraining their effectiveness in tasks that require precise manipulation. 
Although tactile sensing provides direct contact information, it faces unique challenges due to the wide variety of available sensor types. For example, camera-based tactile sensors~\cite{digit, gelsight} are highly heterogeneous, making it difficult to standardize the tactile perception results~\cite{t3}, while magnetic-based tactile sensors~\cite{reskin, tito2018a} often encounter inconsistency issues during replacements~\cite{anyskin}, adding further complexity to their use.

In contact-rich manipulation, integrating force/torque sensing offers an intuitive and versatile approach by directly capturing the physical interactions between the robot and its environment. Since contact inherently produces forces and torques, leveraging this information allows policies to sense and adapt to contact interactions in real time, thereby enhancing the precision and control of manipulation tasks. 
While prior studies~\cite{acp, liu2024force, tacdiffusion} have improved contact-rich task performance by incorporating the force/torque modality, they often \textit{combine force/torque data with vision data through the whole manipulation process, ignoring the fact that force/torque are sparsely activated}. In practice, tasks like wiping involve multiple phases, such as picking up an eraser, performing the wiping, and placing the eraser down. Among these phases, only the wiping phase requires significant contact interactions. During non-contact phases of the task, the inherent noise in force/torque data from real-world sensors might degrade policy performance.

This paper introduces FoAR, a force-aware reactive policy designed for contact-rich robotic manipulation tasks. Building on the state-of-the-art real-world robot imitation policy RISE~\cite{rise}, FoAR effectively integrates high-frequency force/torque sensing with visual inputs by dynamically balancing the usage of force/torque data. This enables precise handling of complex contact dynamics while maintaining strong performance in non-contact phases. The co-design of the FoAR policy and its reactive control strategy further enhances its contact-rich task performance through simple position control. With only 50 demonstrations per task, FoAR significantly outperforms baselines across various challenging contact-rich manipulation tasks. Additionally, FoAR demonstrates exceptional robustness, maintaining stable performance in three evaluation scenarios with unexpected dynamic disturbances, highlighting its adaptability and resilience in real-world applications.

%% file: 2_related_works.tex
\section{Related Works}

\subsection{Integrating Force/Torque Perception in Manipulation}

Force/torque perception is critical for enabling robots to interact effectively with the environment, particularly in manipulations that demand precise control and accurate feedback. By measuring the forces and torques applied to the robot, sensors offer valuable data on contact states, helping the robot perform delicate, contact-rich manipulations~\cite{ftsensor_review}.

Early research leveraged force/torque feedback for low-level control strategies~\cite{Beltran2020learning, hou2019robust, magrini2015control}, enabling precise control in contact-rich tasks but often overlooking its potential for high-level decision-making. More recently, advancements have broadened the application of force/torque perception in robot learning. For example, methods such as~\cite{aburub2024learning, acp, kamijo2024learning} enhance vision-based policies~\cite{diffusionpolicy, act} by incorporating force/torque inputs and predefined stiffness outputs for compliance control. Others have extended the diffusion policy~\cite{diffusionpolicy} into the force domain to predict contact wrenches for hybrid force/position control~\cite{liu2024force}, feedforward forces for impedance control~\cite{tacdiffusion}, and desired forces for admittance control~\cite{zhou2024admittance}.

However, these approaches often only emphasize contact phases by assuming the object is already grasped~\cite{aburub2024learning, kamijo2024learning, tacdiffusion, zhou2024admittance} or fixed to the robot~\cite{acp, liu2024force}, bypassing the impact of noisy force/torque readings during non-contact phases. Other works~\cite{buamanee2024bi, kobayashi2024alpha} employ torque data for bilateral control but rely on leader-follower teleoperation frameworks~\cite{act}, limiting their adaptability to different data collection setups. 

\subsection{Contact-Rich Robotic Manipulation}

Contact-rich manipulation has been extensively studied due to its relevance in both manufacturing and daily life. It involves enabling robots to perform complex tasks that require precise control during physical interactions with the environment~\cite{suomalainen2022survey}. In the past, researchers developed classical force control methods~\cite{hogan1985impedance, mason1981compliance, raibert1981hybrid, whitney1985} for assembly tasks, laying the foundation for contact-rich manipulation control techniques. However, these approaches are often limited by their reliance on precise models and predefined strategies.

Recent advances in learning-based methods have greatly expanded robots' capabilities in contact-rich manipulation. Reinforcement learning-based approaches~\cite{kalakrishnan2011learning, makingsensevisiontouch, levine2015learning, noseworthy2024forge, mimictouch} enable robots to learn complex tasks through interaction, but they often struggle with sim-to-real transfer due to discrepancies in visual observations and force/torque feedback, limiting their performance in real-world tasks. Several imitation learning studies seek to improve the abilities of the robot in contact-rich manipulation abilities by incorporating auxiliary modalities like audio~\cite{playtothescore, seehearfeel, maniwav, hearingtouch}, tactile~\cite{denseboxpacking, playtothescore, huang20243d, seehearfeel, eyesight_hand}, and force/torque~\cite{buamanee2024bi, acp, kamijo2024learning, liu2024force, tacdiffusion, zhou2024admittance}. A key challenge in multimodal policies lies in effectively processing and integrating different modalities within the policy framework, ensuring that the information from each modality is applied to the relevant task phases.

By leveraging the force/torque modality, we propose learning a future contact probability to guide the multimodal feature fusion. This approach allows the force/torque information to enhance the contact phases of the task while preventing noisy data from interfering with other phases, leading to improved performance in various contact-rich manipulation tasks compared to previous fusion methods.

%% file: 3_method.tex
\section{Method}

\if 0
\begin{figure*}
\todo{make this figure much smaller}
    \centering
    \includegraphics[width=\textwidth]{images/contact.pdf}
    \caption{\textbf{Contact State Determination Analysis.} We visualize one demonstration for each task in Fig.~\ref{fig:tasks}, illustrating that visual observation alone - whether using RGB images or point clouds - is insufficient to accurately determine the contact states between the robot and the objects. In contrast, force/torque data (red - $F_x$, green - $F_y$, blue - $F_z$) offers valuable insights into the contact states.}
    \label{fig:contact}
\end{figure*}
\fi

\begin{figure*}
\centering
\begin{minipage}[t]{0.48\textwidth}
\vspace{0pt}
    \centering
    \includegraphics[width=\textwidth]{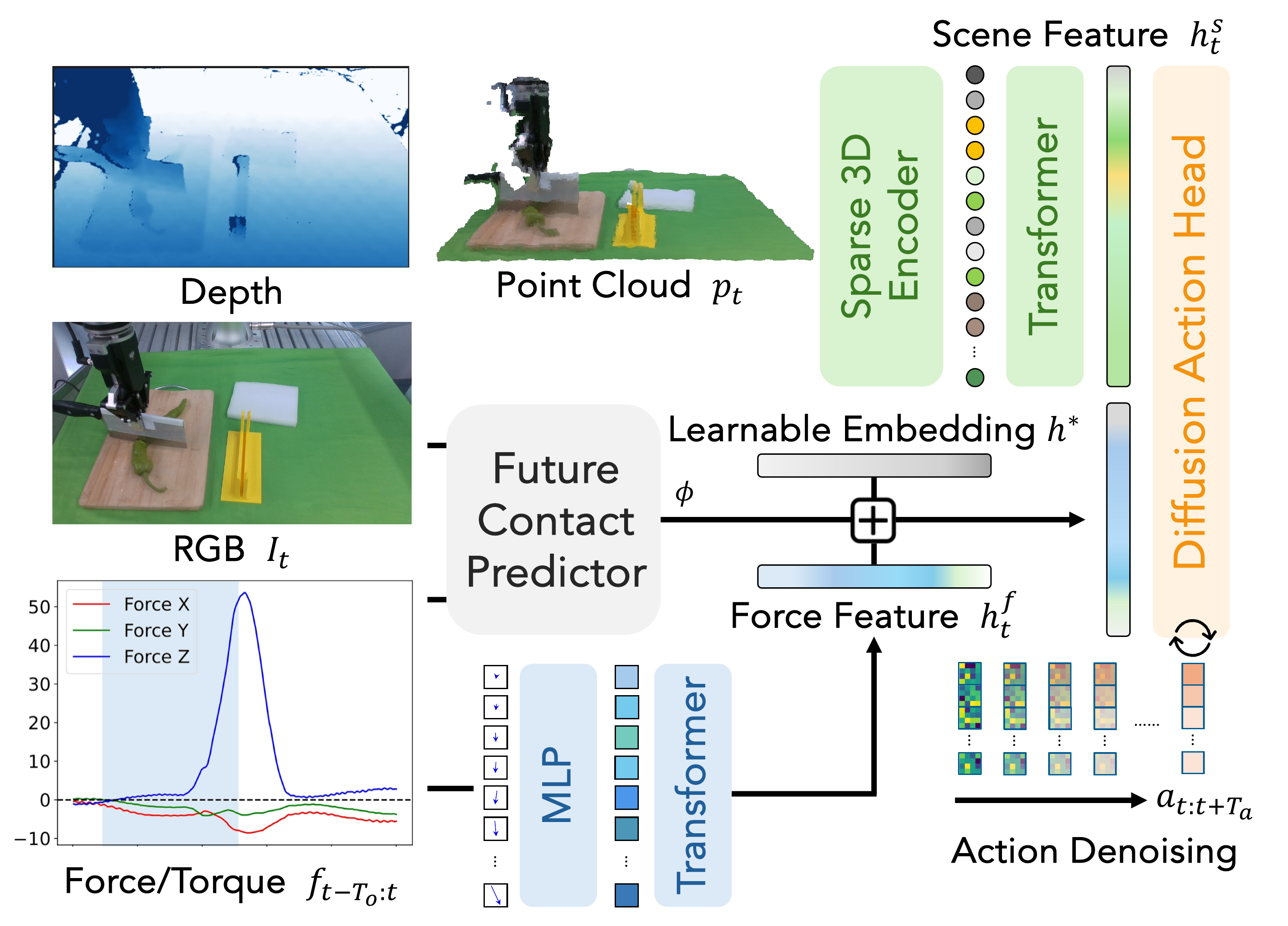}
    \caption{\textbf{FoAR Architecture.} FoAR consists of a point cloud encoder~\cite{rise}, a force/torque encoder, a future contact predictor, and a diffusion action head~\cite{diffusionpolicy}. The scene features and force features are fused under the guidance of the future contact predictor.}
    \label{fig:policy}
\end{minipage}
\hfill
\begin{minipage}[t]{0.48\textwidth}
\vspace{-0.4cm}
\centering
\captionsetup[algorithm]{font=small}
\begin{algorithm}[H]
    \caption{FoAR Inference with Reactive Control}
    \small
    \label{alg:FoAR}
    \begin{algorithmic}[1]
        \State $\textbf{\textit{buffer}}.\text{clear}();$
        \State $\textbf{\textit{contact\_buffer}}.\text{clear}();$ \Comment{\textcolor{darkgrey}{clear the temporal ensemble buffer.}}
        \For{timestep $t \gets 0$ to $N_\text{max} - 1$}
            \If{$t \text{ mod } N_\text{inference} = 0$} \Comment{\textcolor{darkgrey}{at the inference step.}}
                \State $p_t, I_t, f_{t-T_o:t}, q_t \leftarrow \textbf{\textit{agent}}.\text{perception};$\Comment{\textcolor{darkgrey}{perception.}}
                \State $\phi, a_{t:t+T_a} \gets \text{FoAR}(p_t, f_{t-T_o:t}, I_t);$  \Comment{\textcolor{darkgrey}{inference.}}
                \If{$\phi < \delta_\phi$} \Comment{\textcolor{darkgrey}{non-contact phase.}}
                    \State $\textbf{\textit{buffer}}.\text{add}(a_{t:t+T_a});$
                \Else \Comment{\textcolor{darkgrey}{contact phase.}}
                    \If{$\text{force}(f_t) < \delta_f$ and $\text{torque}(f_t) < \delta_t$}
                        \State \Comment{\textcolor{darkgrey}{insufficient force/torque detected.}}
                        \State $\mathbf{d} \gets \text{avg}(a_{t:t+T_f}).\text{pos} - q_t.\text{pos};$
                        \State $a_{t:t+T_a}.\text{pos} \gets a_{t:t+T_a}.\text{pos} + \epsilon \cdot \mathbf{d} / \|\mathbf{d}\|_2;$
                        \State \Comment{\textcolor{darkgrey}{update actions towards predicted direction.}}
                    \EndIf
                    \State $\textbf{\textit{contact\_buffer}}.\text{add}(a_{t:t+T_a});$
                \EndIf
            \EndIf
            \State $a_t \gets \textbf{\textit{buffer}}.\text{get}(t) \textbf{ if } \phi < \delta_\phi \textbf{ else } \textbf{\textit{contact\_buffer}}.\text{get}(t);$
            \State $\textbf{\textit{agent}}.\text{execute}(a_t);$  \Comment{\textcolor{darkgrey}{retrieve and execute the action.}}
        \EndFor
    \end{algorithmic}
\end{algorithm}
\end{minipage}
\vspace{-0.3cm}
\end{figure*}

\subsection{Preliminary}\label{sec:method-preliminary}

Given an observation $p_t\in\mathbb{R}^{N_t \times 6}$ \red{with $N_t$ points extracted from RGB-D image} at current timestep $t$, RISE~\cite{rise} $\pi(p_t) = a_{t:t+T_a}$ learns a direct mapping from the current observation to future robot actions over a horizon of $T_a$. Building upon RISE, our proposed force-aware policy, FoAR, incorporates high-frequency force/torque observations $f_{t-T_o:t}\in\mathbb{R}^{T_o\times 6}$ over a historical horizon of $T_o$ as additional inputs. Notice that $T_o$ represents the history horizon for high-frequency force/torque data, typically sampled at about 100Hz, while $T_a$ denotes the action horizon for future predictions, which operates at a lower frequency like 10Hz. 

While vision-based policies~\cite{diffusionpolicy, octo, rise, cage} have demonstrated success in simple contact-rich tasks, we argue that force/torque information is vital for more complex scenarios. Taking the determination of contact states as an example, Fig.~\ref{fig:teaser} (left) shows that visual differences in either RGB images or point clouds before and after contact are minimal, making it difficult to determine contact states. In contrast, force/torque data provides clear and reliable indicators of contact, highlighting its critical role in such tasks. As a result, incorporating high-frequency force/torque data complements point cloud observations, enabling more accurate and robust decision-making in contact-rich manipulations.

\subsection{Force-Aware Policy Design}

\textbf{Point Cloud Encoder.} Following RISE~\cite{rise}, we employ sparse 3D encoder~\cite{minkowski} with a shallow ResNet architecture~\cite{resnet} to process the point cloud $p_t\in\mathbb{R}^{N_t\times 6}$ into sparse point tokens $P_t\in\mathbb{R}^{N_p\times 512}$, \red{where $N_p$ represents the number of sparse point tokens after processing.} A Transformer~\cite{transformer} with sparse point encodings~\cite{rise} is then applied to these point tokens to generate a scene feature $h^s_t \in \mathbb{R}^{512}$.

\textbf{Force/Torque Encoder.} The force/torque observation $f_t\in\mathbb{R}^6$ is first processed through a 3-layer MLP to generate the corresponding force token $F_t\in\mathbb{R}^{512}$. These tokens over the past horizon $F_{t-T_o:t}\in\mathbb{R}^{T_o\times 512}$, being inherently time-series data in nature, are then encoded using a Transformer~\cite{transformer} with sinusoidal positional encodings applied along the temporal axis, resulting in a force feature $h^f_t \in \mathbb{R}^{512}$.

\textbf{Feature Fusion.} Previous studies on multimodal feature fusion in robotic manipulation have explored approaches such as direct concatenation of features~\cite{makingsensevisiontouch, liu2024force} and processing multimodal tokens through Transformers~\cite{vtt, seehearfeel, maniwav, octo}. However, such simple fusion methods often lead to the noisy force modality interfering with the non-contact phases of the task. Instead, we introduce a future contact predictor $\phi(t) \in [0,1]$ to guide the feature fusion process. Specifically, the fused feature $h_t$ is calculated as follows:
$$
    h_t = \left[h^s_t; \phi(t) \cdot h^f_t + (1- \phi(t)) \cdot h^*\right],
$$
where $h^*$ is a learnable embedding, and $[\cdot;\cdot]$ is the concatenation symbol. In other words, the future contact predictor dynamically adjusts the weight of the force feature $h_t^f$ in the fusion process, ensuring that the force data is strongly emphasized during contact phases while minimizing its impact during non-contact phases by blending it with a neutral embedding $h^*$. This approach allows the policy to more effectively utilize multimodal information without introducing interference from irrelevant data.

\textbf{Future Contact Predictor.} The future contact predictor takes the current observations as inputs and outputs the probability that contact will occur in the \textit{future} steps. This probability is used to modulate the fusion of the visual and force modalities, allowing the model to emphasize force data when contact is likely to occur and reduce its influence during non-contact phases. As discussed in \S\ref{sec:method-preliminary}, we use current RGB image $I_t$ and force/torque data $f_{t-T_o:t}$ as observation inputs to the predictor, since (1) using RGB images can make the predictor more lightweight given that it performs similarly with point clouds in contact state determination; (2) while force/torque data does not directly predict future contact, it helps correct the predictor when unexpected contact occurs.

\textbf{Action Head.} The fused feature $h_t$ is then used as the conditioning input for the action denoising process~\cite{diffusionpolicy,ddpm,ddim} to generate robot end-effector actions by progressively refining noisy action trajectories.

\begin{figure*}
    \centering
    \includegraphics[width=\textwidth]{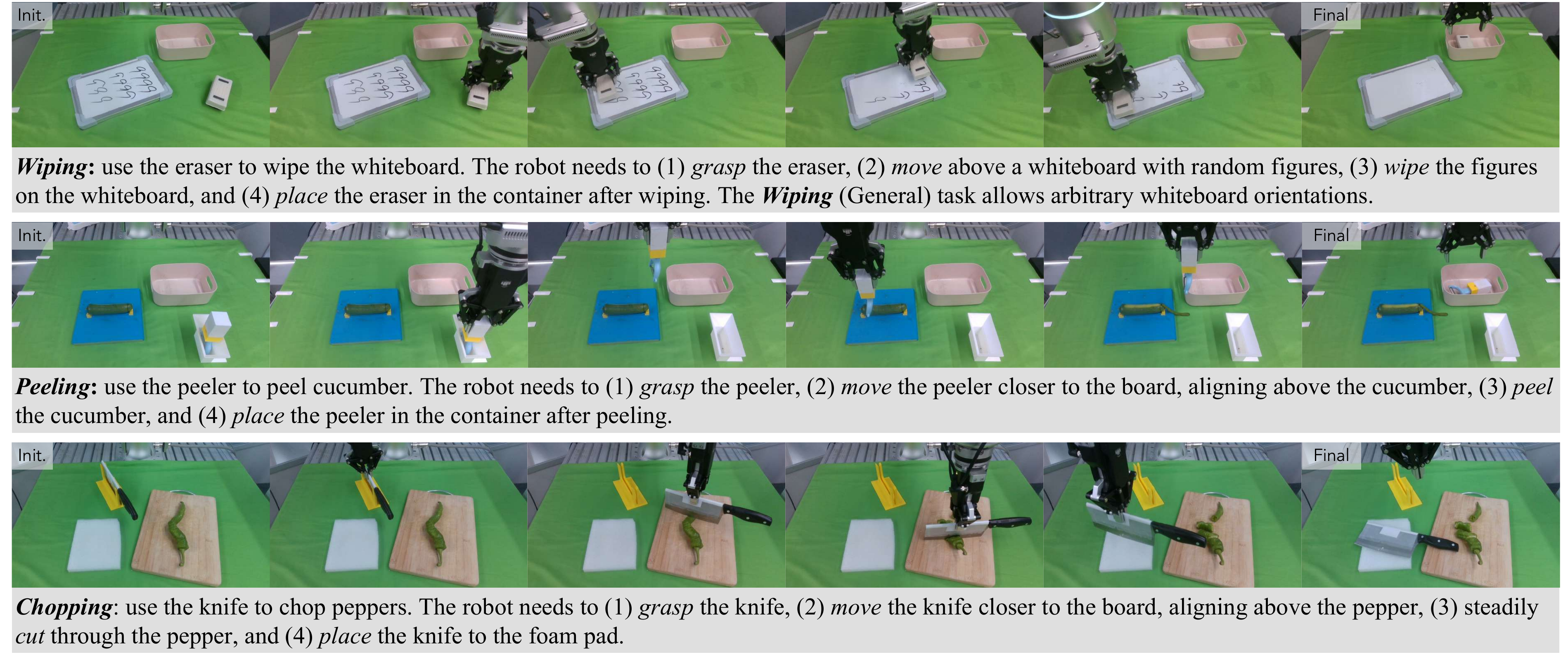}
    \caption{\textbf{Tasks}. We carefully design 3 challenging contact-rich tasks that focus on different aspects of the contact-rich manipulations. These tasks involve both non-contact phases and contact phases to evaluate the policy performance thoroughly.}
    \label{fig:tasks}\vspace{-0.3cm}
\end{figure*}

\textbf{Supervision.} The generated action is supervised by ground-truth action in demonstration data via L2 loss $\mathcal{L}_\text{action}$ in the diffusion process. The ground-truth future contact state is automatically extracted from the demonstrations based on whether the force/torque data exceeds a threshold $\delta_\text{demo}$ within a surrounding time window around the current timestep, which supervises the future contact predictor through binary cross-entropy loss $\mathcal{L}_\text{predictor}$. The overall loss $\mathcal{L}$ is a linear combination of both terms:
$$
\mathcal{L} = \mathcal{L}_\text{action} + \alpha \mathcal{L}_\text{predictor},
$$
where $\alpha$ is the weighting factor.

\subsection{Reactive Control in Deployment}

%todo: add more refs
%compliance control~\cite{acp}, hybrid force/position control~\cite{liu2024force}, admittance control~\cite{}
Prior literature has explored various control strategies for contact-rich manipulation, such as admittance control~\cite{zhou2024admittance}, compliance control~\cite{acp, kamijo2024learning, mason1981compliance}, and hybrid force/position control~\cite{liu2024force, raibert1981hybrid}. These approaches often require additional parameters, such as stiffness and contact force direction. In contrast, we demonstrate that our proposed future contact predictor enables the robot to perform accurate, force-feedback-driven manipulation in contact-rich tasks even using simple end-effector position control, eliminating the need for complex parameter tuning and prediction.

We introduce reactive control during deployment, as outlined in Alg.~\ref{alg:FoAR}. Specifically, we threshold the predicted future contact probability $\phi$ from the contact predictor to determine whether the robot will make contact with the object and whether the predicted end-effector action needs to be adjusted using force/torque feedback. If $\phi$ exceeds the threshold $\delta_\phi$, indicating that the robot is in contact or will soon make contact with the object, the controller will check the current force/torque readings $f_t$, and correct the predicted robot actions if insufficient force/torque is detected. For action correction (Line 12-14 in Alg.~\ref{alg:FoAR}), we estimate the future action direction based on the predicted action chunk and the current end-effector pose $q_t$, then adjust the predicted robot actions by a small step $\epsilon$ towards that direction. Different temporal ensemble buffers~\cite{act} are used for contact and non-contact phases to avoid mutual interference while ensuring smooth trajectory execution.

By incorporating reactive control during deployment, our FoAR policy can effectively handle uncertainties and dynamic changes in the environment, allowing the robot to adapt to real-world variations and achieve more reliable contact-rich manipulation performance.

%% file: 4_experiments.tex
\section{Experiments}

During the experiments, we intend to answer the following research questions:
\begin{itemize}
    \item \textbf{\textit{Q1}}: Does integrating force/torque information improve policy performance and manipulation accuracy in contact-rich tasks, particularly in real-world scenarios where such tasks involve multiple phases with varying demands on precision and contact interactions?
    \item \textbf{\textit{Q2}}: Is the feature fusion module of the FoAR policy more effective than other variants in terms of integrating force/torque information?
    \item \textbf{\textit{Q3}}: \red{Can contact predictor, reactive control, and high-frequency force/torque sensing enhance the policy's ability to perform contact-rich operations?}
    \item \textbf{\textit{Q4}}: Can FoAR maintain consistent task performance under unexpected environmental disturbances?
\end{itemize}

\subsection{Setup}

\begin{figure*}
\centering
\begin{minipage}{0.6\textwidth}
    \centering
    \footnotesize
    \setlength\tabcolsep{2.5pt}
    \begin{tabular}{ccccccccccc}    
    \toprule
    \multirow{3}{*}{\textbf{Method}} & \multicolumn{3}{c}{\textbf{\textit{Wiping}}} & \multicolumn{3}{c}{\textbf{\textit{Wiping}} (General)} & \multicolumn{3}{c}{\textbf{\textit{Peeling}}} \\
    \cmidrule(lr){2-4}
    \cmidrule(lr){5-7}
    \cmidrule(lr){8-10}
    & \multirow{2}{*}{\textbf{Score} $\uparrow$} & \multicolumn{2}{c}{\textbf{ASR} (\%) $\uparrow$} & \multirow{2}{*}{\textbf{Score} $\uparrow$} & \multicolumn{2}{c}{\textbf{ASR} (\%) $\uparrow$} & \multirow{2}{*}{\textbf{Score} $\uparrow$} & \multicolumn{2}{c}{\textbf{ASR} (\%) $\uparrow$} \\ \cmidrule(lr){3-4}\cmidrule(lr){6-7} \cmidrule(lr){9-10}
    & & Grasp & Wipe & & Grasp & Wipe & & Grasp & Peel \\ 
    \midrule
    \red{ACT~\cite{act}} & \red{0.275} & \red{65} & \red{50} & \red{0.250} & \red{65} & \red{50}  & \red{0.120} & \red{95} & \red{25}\\
    \red{Diffusion Policy~\cite{diffusionpolicy}} & \red{0.400} & \red{75} & \red{60} & \red{0.350} & \red{75} & \red{50} & \red{0.386} & \red{85} & \red{70} \\
    RISE~\cite{rise} & 0.500 & \textbf{100} & 75 & 0.500 & 90 & 80 & 0.377 & \textbf{100} & 50 \\ 
    \midrule
    RISE (force-token) & 0.575 & 85 & 80 & 0.600 & 90 & 80 & 0.487 & 95 & 75\\
    RISE (force-concat) & 0.475 & \textbf{100} & 65 & \red{0.675} & \red{\textbf{100}} & \red{95} & 0.524 & \textbf{100} & 80 \\
    FoAR (3D-cls) & 0.175 & 40 & 35 & \red{0.200} & \red{40} & \red{40} & 0.270 & 95 & 40\\
    % FoAR (force-token) & \\
    % FoAR (force-concat) & \\
    % FoAR (force-binary) & \\
    % FoAR (force-no-predictor) & 0.650 & \textbf{100} & 85 \\
    % FoAR (2 Hz) & 0.625 & \textbf{100} & 85\\
    % FoAR (10 Hz) & 0.800 & \textbf{100} & \textbf{100}\\
    
    \midrule
    FoAR (\textit{ours}) & \cellcolor[HTML]{CAD4E7}0.875 & \textbf{100} & \textbf{100} & \cellcolor[HTML]{CAD4E7}0.850 & \textbf{100} & \textbf{100} & \cellcolor[HTML]{CAD4E7}0.756 & \textbf{100} & \textbf{100}\\
    \bottomrule
    \end{tabular}
    \captionof{table}{\textbf{Evaluation Results of the \textit{Wiping} and \textit{Peeling} Tasks.} ASR denotes the action success rate, measuring the success rates of the robot in executing certain actions, regardless of the quality of the actions. } \label{tab:wipe-peel}
\end{minipage}
\hfill
\begin{minipage}{0.38\textwidth}
        \centering
        \includegraphics[width=0.75\textwidth]{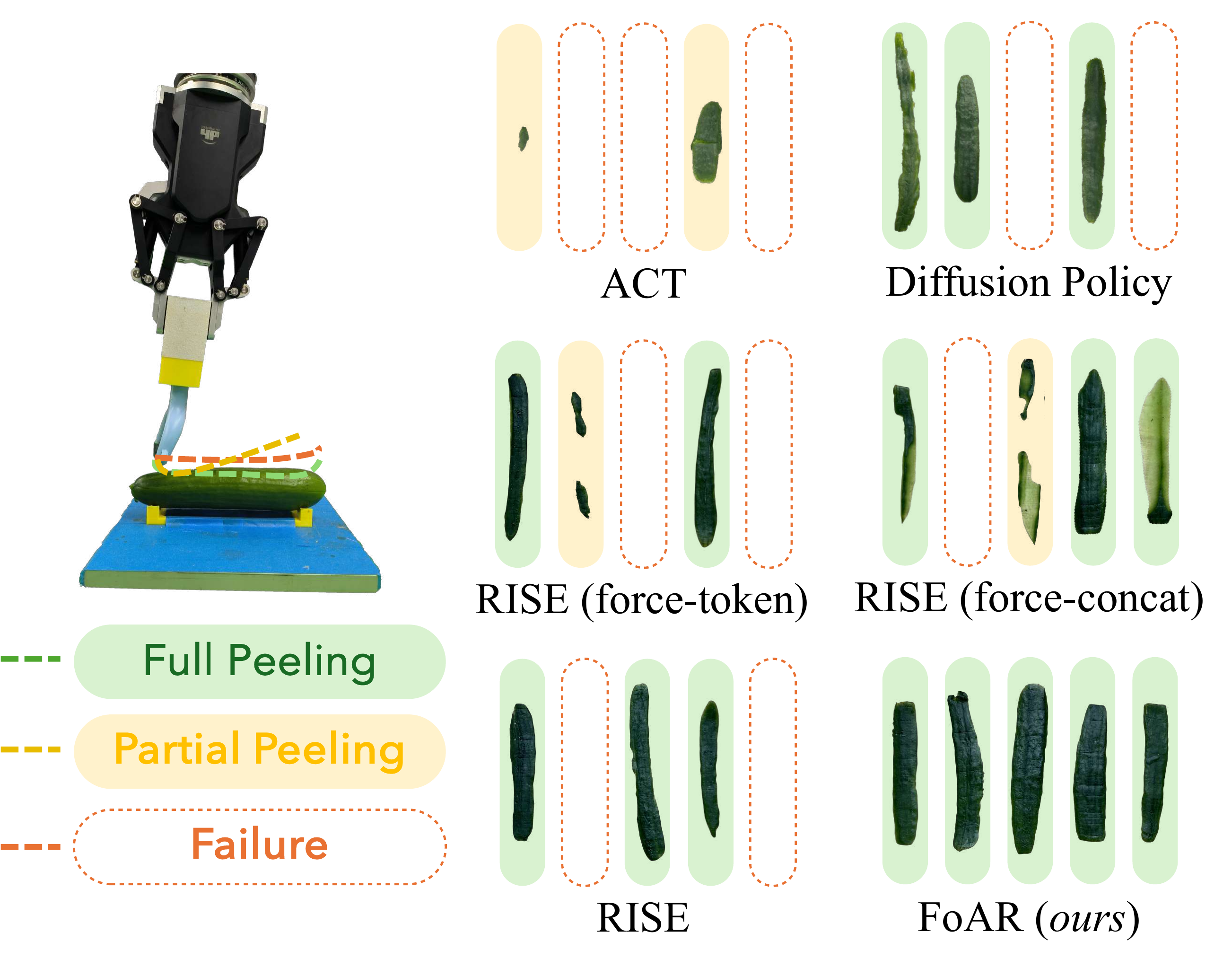}
        \captionof{figure}{\textbf{Qualitative Results of the \textit{Peeling} Task}. \red{The green background indicates the robot fully peels the cucumber, while yellow and red backgrounds represent partial peeling and peeling failure, respectively.}}
        \label{fig:peel}
\end{minipage}
\vspace{-0.3cm}
\end{figure*}

\textbf{Platform.} The experimental platform consists of a Flexiv Rizon robotic arm with a Dahuan AG-95 gripper, and an OptoForce force/torque sensor mounted between the flange and the gripper. The robot operates within a 45cm $\times$ 60cm $\times$ 40cm workspace. An Intel RealSense D435 RGBD camera located in front of the robot workspace is used for scene perception. All devices are linked to a workstation with an Intel Core i9-10900K CPU and an NVIDIA RTX 3090 GPU for both data collection and evaluation.

\textbf{Tasks.} As shown in Fig.~\ref{fig:tasks}, we design three challenging contact-rich tasks across two categories: surface force control (\textbf{\textit{Wiping}} and \textbf{\textit{Peeling}}) and instantaneous force impact (\textbf{\textit{Chopping}}). These tasks require different capabilities in terms of the direction, intensity, and precision of applied contact forces. Moreover, these tasks are designed to have both non-contact phases and contact phases for thorough evaluations. For the \textbf{\textit{Wiping}} task, we design two variants: one with a fixed orientation of the whiteboard, and another that allows arbitrary orientations, denoted as \textbf{\textit{Wiping}} (General).

\textbf{Baselines.} We evaluate our proposed approach against six baseline methods, including the vision-based policy ACT \cite{act}, Diffusion Policy \cite{diffusionpolicy}, and RISE \cite{rise}, as well as three ablation variants: (1) \textit{RISE (force-token)}: incorporates encoded force/torque information as additional tokens within the RISE transformer, akin to \cite{vtt, seehearfeel, maniwav, octo}; (2) \textit{RISE (force-concat)}: directly concatenates the force feature with the vision feature for action generation; (3) \textit{FoAR (3D-cls)}: uses scene features $h_t^s$ directly in the future contact predictor, instead of a separate image encoder. 

\begin{figure}[t]
    \centering
    \includegraphics[width=0.9\linewidth]{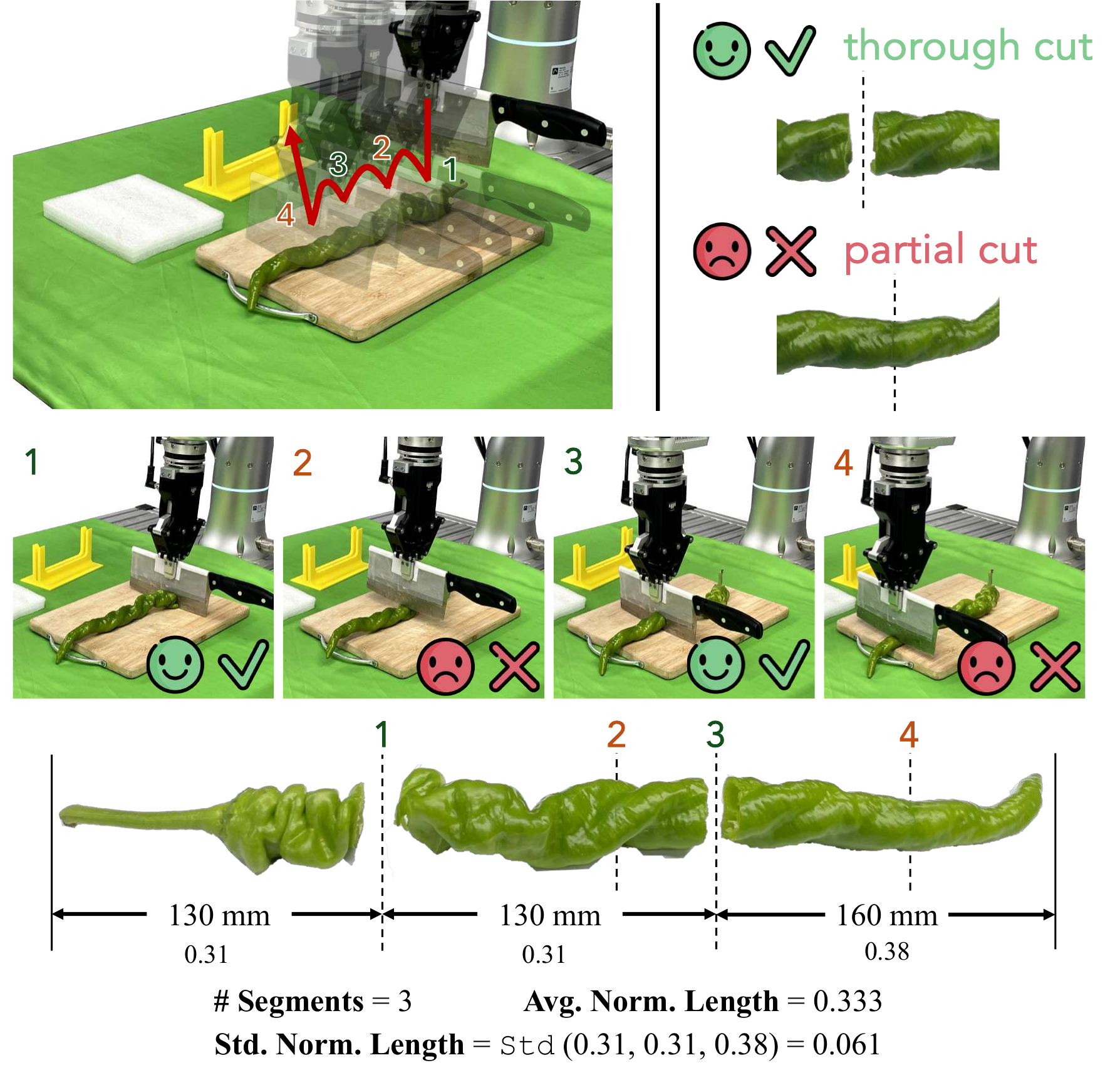}
    \caption{\textbf{Evaluation Metrics of the \textit{Chopping} Task}. We encourage the robot to divide the pepper into several uniform small segments, without segments sticking together due to partial cuts.}
    \label{fig:chop-metric}
\end{figure}

\textbf{Metrics.} For all tasks, we calculate the action success rate (referred to as ASR) to assess the policy's ability to meet basic action requirements, regardless of action quality. For the \textbf{\textit{Wiping}} task, the score is assigned to 1 for a fully wiped whiteboard, 0.5 for partial wiping, and 0 for no erasure. For the \textbf{\textit{Peeling}} task, the score is calculated based on the proportion of peeled cucumber skin to the total cucumber length, normalized by the average proportion in the demonstration data (0.778). 
For the \textbf{\textit{Chopping}} task, we aim to let the robot use the knife to divide the pepper into several uniform small segments. Therefore, we focus on the number of segments, as well as the mean and standard deviation of the normalized lengths (defined as the proportion of each segment’s length to the total length of the pepper), providing a comprehensive assessment of chopping precision and consistency, as shown in Fig. \ref{fig:chop-metric}.

\textbf{Protocols.} For policy training, we collect 50 expert demonstrations for the \textbf{\textit{Wiping}} and \textbf{\textit{Peeling}} tasks, and 40 for the \textbf{\textit{Chopping}} task through haptic teleoperation \cite{rh20t}. During evaluation, we run 20 trials per method for the \textbf{\textit{Wiping}} and \textbf{\textit{Peeling}} tasks, and 10 trials each only for FoAR and RISE~\cite{rise} on the \textbf{\textit{Chopping}} task to conserve resources. Objects are randomly placed in the workspace, while ensuring similar positions across methods for fair comparisons.

\textbf{Implementation.}
FoAR uses $T_o = 200$ to encode high-frequency (100Hz) force/torque data, corresponding to approximately 2 seconds of data. The dimensions of force tokens, scene feature $h_t^s$, force feature $h_t^f$, and learnable embedding $h^*$ are all set to 512. For the future contact predictor, we utilize a ResNet18~\cite{resnet} vision encoder and an MLP-based force encoder, followed by feature concatenation and a linear layer to output the probability $\phi$. We combine the action loss and the predictor loss using $\alpha = 0.1$ during training. Other hyperparameters remain the same as RISE. For reactive control in deployment, we set the future contact probability threshold $\delta_\phi = 0.9$, force threshold $\delta_f = 8\text{N}$, torque threshold $\delta_t = 5\text{N}\cdot\text{m}$, and small step $\epsilon = 0.006\text{m}$. 

\subsection{Surface Force Control Tasks: \textbf{\textit{Wiping}} and \textbf{\textit{Peeling}}}

In surface force control tasks (\textbf{\textit{Wiping}} and \textbf{\textit{Peeling}}), the robot utilizes force/torque data to maintain consistent surface contact. A key challenge arises from the variability in tool grasp positions (\textit{e.g.}, top, bottom, center, or off-center), requiring adaptive control to adjust for changes in the grasp. As shown in Fig. \ref{fig:tasks}, the \textbf{\textit{Wiping}} task assesses the ability of the policy to maintain continuous and sustained contact, while the \textbf{\textit{Peeling}} task emphasizes precision and sensitivity in manipulation.

\textbf{FoAR significantly outperforms baselines in surface force control tasks by integrating force/torque information to enhance manipulation accuracy and contact consistency in surface force control tasks (\textit{Q1}).} 
We report the evaluation results for the \textbf{\textit{Wiping}}, \textbf{\textit{Wiping}} (General), and \textbf{\textit{Peeling}} tasks in Table~\ref{tab:wipe-peel}. Our FoAR policy achieves the highest scores of 0.875, 0.850, and 0.756 for the \textbf{\textit{Wiping}}, \textbf{\textit{Wiping}} (General), and \textbf{\textit{Peeling}} tasks, respectively, significantly outperforming all baselines and variants. FoAR attains 100\% success rates in both grasping the tool and performing contact-rich operations (wiping and peeling) in all tasks, demonstrating its ability to maintain continuous and precise contact regardless of the grasp position of the tool (eraser and peeler). In contrast, the pure vision-based policy RISE struggles with these contact-rich operations due to the lack of force/torque feedback, leading to inaccurate position control, which reflects the difficulty in maintaining consistent contact stemming from the absence of force/torque perceptions. The qualitative results of the \textbf{\textit{Peeling}} task in Fig.~\ref{fig:peel} further support these findings, showcasing that FoAR achieves more consistent and effective performance compared to the baselines, which often result in partial peelings or failures.

\textbf{The feature fusion module of FoAR enhances capabilities in contact-rich operations while maintaining strong performance during non-contact phases, surpassing several variants in integrating force/torque information (\textit{Q2}).} We report the evaluation results for these variants in Table~\ref{tab:wipe-peel} and Fig.~\ref{fig:peel}. RISE (force-token) and RISE (force-concat) exhibit similar or slightly better performance compared to the pure vision-based policy RISE, suggesting that incorporating force/torque data as an additional input does provide some benefits. However, the key factor lies in how these inputs are effectively leveraged in the policy. Simply integrating force/torque tokens into the policy transformer or concatenating force features with vision features not only fails to fully leverage force/torque information but also risks introducing noisy force/torque data during non-contact phases, which can interfere with the policy’s decision-making process and thus negatively impacting overall performance, \textit{e.g.}, leading to lower grasp action success rates in both tasks for RISE (force-token). On the contrary, FoAR demonstrates strong performance in both contact and non-contact phases, highlighting the effectiveness of our feature fusion module over these variants in utilizing force/torque data.

\begin{table}[t]
    \centering
    \setlength\tabcolsep{2.5pt}
    \begin{tabular}{cccccc} 
    \toprule
    \multirow{2}{*}{\textbf{Method}}       & \multirow{2}{*}{\textbf{\# Segments} $\uparrow$}  & \multicolumn{2}{c}{\textbf{Norm. Length}} & \multicolumn{2}{c}{\textbf{ASR} (\%) $\uparrow$} \\
    \cmidrule(lr){3-4}\cmidrule(lr){5-6}
    & & \textbf{Avg.} $\downarrow$ & \textbf{Std.} $\downarrow$ & Grasp & Place\\ 
    \midrule
    RISE~\cite{rise} & 1.8 $\pm$ 0.6 & 0.727      & 0.411 & \textbf{100} & 30 \\ 
    FoAR~(\textit{ours}) & \textbf{3.9} $\pm$ 0.9  & \textbf{0.353}       & \textbf{0.094} & \textbf{100} & \textbf{70}  \\
    \midrule
    {\color{darkgrey}\textit{Oracle} (demonstration)} & {\color{darkgrey}5.0 $\pm$ 0.0} & {\color{darkgrey}0.200} & {\color{darkgrey}0.056} & {\color{darkgrey} 100} & {\color{darkgrey} 100}\\
    \bottomrule
    \end{tabular}
    \caption{\textbf{Evaluation Results of the \textit{Chopping} Task.} We also calculate the metrics of the demonstrations as an oracle for reference.}
    \label{tab:res-chop}\vspace{-0.4cm}
\end{table}

\textbf{Separating the future contact predictor from the policy backbone is crucial to avoid disruption (\textit{Q2}).} As shown in Table~\ref{tab:wipe-peel}, the FoAR (3D-cls) variant even significantly underperforms the RISE baseline. This variant employs a shared sparse 3D encoder for both the future contact predictor and the policy backbone. We suspect that the visual features required by each component differ substantially. For example, in the \textbf{\textit{Wiping}} task, the future contact predictor focuses on whether the eraser is positioned above the whiteboard, whereas the policy requires detailed information like the precise locations of objects and the end-effector position. Consequently, sharing a single vision encoder may cause conflicting attention and potential interference, disrupting both components and reducing their effectiveness.

\subsection{Instantaneous Force Impact Task: \textbf{\textit{Chopping}}}

The \textbf{\textit{Chopping}} task evaluates the robot's ability to handle instantaneous force impacts, requiring precise force and torque control that vision data alone cannot provide~\cite{cage}. 
The main challenge lies in accurately assessing the chopping as the knife's contact with the pepper and the chopping depth constantly change.

\textbf{Vision alone is insufficient for ensuring smooth chops, highlighting the necessity of force/torque feedback for precise control and improved policy performance (\textit{Q1}).}
The results in Tab.~\ref{tab:res-chop} demonstrate that FoAR outperforms the baseline policy RISE, providing more reliable and controlled performance in the \textbf{\textit{Chopping}} task. It achieves over double the number of segments (3.9 \textit{v.s.} 1.8) and a lower averaged normalized segment length (0.353 \textit{v.s.} 0.727) with reduced segment variability (standard deviation of 0.094 \textit{v.s.} 0.411), indicating better performance in chopping the pepper into smaller, more uniform segments.

\subsection{Ablations}
\textbf{\red{Designing contact predictor, applying reactive control, and integrating high-frequency force/torque sensing} enable the policy to perform more precise contact-rich manipulations (\textit{Q3})}. To illustrate the importance of these design, we take the \textbf{\textit{Wiping}} task as an example. \red{By comparing with a policy variant using contact \textit{detection} instead of future contact \textit{prediction}, we observe that the detection variant occasionally fails to make contact, indicating that contact detection may lead to delayed or even failed contact.} 
Results in Tab.~\ref{tab:reactive-control} also demonstrate that reactive control is essential for optimal policy performance. Notably, our proposed reactive control relies on the predicted future contact probability from the future contact predictor in the policy, highlighting the effective co-design of the FoAR policy and the reactive control mechanism. 
\red{Additionally, relatively high-frequency (100Hz) force/torque sensing helps policy achieve better performance. Although there is a trade-off between high-frequency force/torque sensing and computational efficiency, we found 100 Hz as a balanced compromise.} % Consequently, in all experiments, reactive control is applied to FoAR-based methods to ensure high performance.

\begin{table}[htbp]    
    \centering
    \footnotesize
    \setlength\tabcolsep{2.5pt}
    \begin{tabular}{cccccc}    
    \toprule
    \multicolumn{3}{c}{\textbf{Design Choice}} & \multirow{2}{*}{\textbf{Score} $\uparrow$} & \multicolumn{2}{c}{\textbf{ASR} (\%) $\uparrow$}
    \\ \cmidrule(lr){1-3}\cmidrule(lr){5-6}
    F/T Freq. & \textit{w.} Predictor & \textit{w.} Reactive & & Grasp & Wipe \\ 
    \midrule    
    \red{100Hz} & & \red{\checkmark} & \red{0.650} & \red{\textbf{100}} & \red{85}  \\
    100Hz & \checkmark & & 0.650 & \textbf{100} & 80 \\
    %\midrule
    \midrule
    \red{2Hz} & \red{\checkmark} & \red{\checkmark} & \red{0.625} & \red{\textbf{100}} & \red{85} \\
    \red{10Hz} & \red{\checkmark} & \red{\checkmark} & \red{0.800} & \red{\textbf{100}} & \red{\textbf{100}} \\
    \midrule
    100Hz & \checkmark & \checkmark & \cellcolor[HTML]{CAD4E7}0.875 & \textbf{100} & \textbf{100} \\
    \bottomrule
    \end{tabular}
    \caption{\textbf{Ablation Results of the \textit{Wiping} Task on Several Design Choices.} \red{We ablate our design choices on contact predictor, reactive control, and high-frequency force/torque sensing.}}
    \label{tab:reactive-control}
\end{table}

\subsection{Robustness to Dynamic Disturbances}

\begin{figure*}
\centering
\begin{minipage}{0.24\textwidth}
        \centering
        \includegraphics[width=0.9\textwidth]{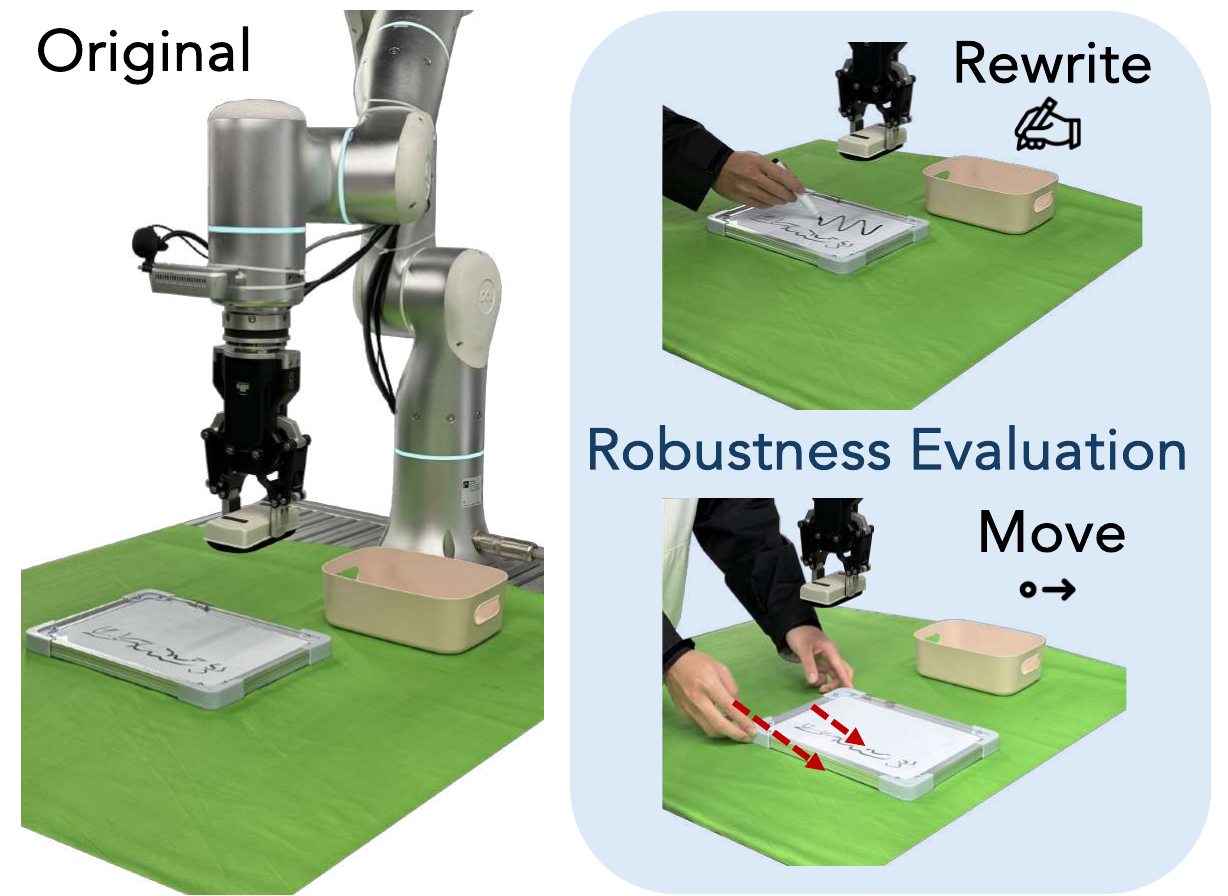}
\end{minipage}
\hfill
\begin{minipage}{0.75\textwidth}
    \centering
    \footnotesize
    \setlength\tabcolsep{2.5pt}
    \begin{tabular}{cccccccccccccc}    
    \toprule
    \multirow{3}{*}{\textbf{Method}} & \multicolumn{3}{c}{\textbf{\textit{Original}}} & \multicolumn{3}{c}{\textbf{\textit{Rewrite}}} & \multicolumn{3}{c}{\textbf{\textit{Move}}} & \multicolumn{3}{c}{\textbf{\textit{Rewrite + Move}}} \\
    \cmidrule(lr){2-4}
    \cmidrule(lr){5-7}
    \cmidrule(lr){8-10}
    \cmidrule(lr){11-13}
    & \multirow{2}{*}{\textbf{Score} $\uparrow$} & \multicolumn{2}{c}{\textbf{ASR}(\%) $\uparrow$} & \multirow{2}{*}{\textbf{Score} $\uparrow$} & \multicolumn{2}{c}{\textbf{ASR}(\%) $\uparrow$} & \multirow{2}{*}{\textbf{Score} $\uparrow$} & \multicolumn{2}{c}{\textbf{ASR}(\%) $\uparrow$}  & \multirow{2}{*}{\textbf{Score} $\uparrow$} & \multicolumn{2}{c}{\textbf{ASR}(\%) $\uparrow$} \\ \cmidrule(lr){3-4}\cmidrule(lr){6-7} \cmidrule(lr){9-10} \cmidrule(lr){12-13}
    & & Grasp & Wipe & & Grasp & Wipe & & Grasp & Wipe & & Grasp & Wipe \\ 
    \midrule
    RISE~\cite{rise} & 0.500 & 90 & 80 & 0.500 & 80 & 70 & 0.600 & \textbf{100} & \textbf{100} & 0.500 & \textbf{100} & 70 \\ 
    RISE (force-token) & 0.600 & 90 & 80 & 0.450 & 90 & 90 & 0.500 & 90 & 80 & 0.600 & \textbf{100} & \textbf{100}\\
    \midrule
    FoAR (\textit{ours}) & \cellcolor[HTML]{CAD4E7}0.850 & \textbf{100} & \textbf{100} & \cellcolor[HTML]{CAD4E7}0.800 & \textbf{100} & \textbf{100} & \cellcolor[HTML]{CAD4E7}0.850 & \textbf{100} & \textbf{100} & \cellcolor[HTML]{CAD4E7}0.800 & \textbf{100} & \textbf{100} \\
    \bottomrule
    \end{tabular}
\end{minipage}
\captionof{table}{\textbf{Robustness Evaluation Results of the \textit{Wiping} (General) Task.} The figure on the left illustrates the dynamic disturbances in the robustness evaluation. ``Original'' refers to vanilla evaluation with no disturbances.} \label{tab:wipe-reactive}\vspace{-0.3cm}
\end{figure*}

To further assess the adaptability of our model FoAR under more challenging and varied conditions, we develop three robustness evaluations for the \textbf{\textit{Wiping}} (General) task: (1) \textit{Rewrite}: write new random figures on the wiped area after robot wiping; (2) \textit{Move}: move the whiteboard to a different position after robot wiping; (3) \textit{Rewrite + Move}: combine the previous two dynamic disturbances, \textit{i.e.}, move the whiteboard to a different position, and write new random figures on the wiped area after robot wiping.
These evaluations introduce disturbances during task execution to assess how well the methods can adjust to new conditions.

\textbf{FoAR maintains consistent task performance under unexpected and dynamic environmental disturbances, demonstrating superior robustness and adaptability (\textit{Q4}).}
As shown in Tab.~\ref{tab:wipe-reactive}, FoAR successfully maintains consistent performance across all robustness evaluations, adapting to several dynamic environmental disturbances. While RISE also demonstrates strong generalization ability~\cite{rise}, its performance is limited by the absence of force/torque feedback. Built upon RISE, FoAR inherits this generalization ability while leveraging force/torque integration to achieve superior performance. In contrast, RISE (force-token) struggles in these complex scenarios, likely due to disturbances forcing the policy into non-contact phases, requiring action re-generation. Noise in force/torque data during these transitions further amplifies errors, hindering its effectiveness.

%% file: 6_conclusions.tex
\section{Conclusion}
In this paper, we propose FoAR, a force-aware reactive policy tailored for contact-rich robotic manipulation. By introducing a future contact predictor, the policy enables effective contact-guided feature fusion between force/torque and visual information, dynamically balancing the contribution of each modality based on future contact probability. This design not only enhances precision during contact phases but also maintains strong performance in non-contact phases. Additionally, the future contact probability further guides the reactive control strategy, improving policy performance even with simple position control. Extensive experiments demonstrate the superior performance of FoAR in contact-rich tasks that require sustained and precise contact, such as wiping, peeling, and chopping. 

\red{While FoAR demonstrates notable progress in contact-rich robotic manipulation, it has several limitations. The use of static force/torque thresholds, while effective in the evaluated tasks, may struggle in complex environments. Additionally, the current policy relies on the simple end-effector position control mechanism, limiting its adaptability. Future work will explore leveraging advanced control strategies, such as compliance control and hybrid force/position control, to enhance performance. Extending the FoAR policy to dual-arm robots, humanoid robots, and dexterous hands is another promising direction for enabling more complex and dexterous contact-rich manipulations.}

%% file: acknowledgement.tex
\section*{Acknowledgement}

We would like to thank Chenxi Wang for helpful discussions, Yiming Wang and Shangning Xia for their help during the data collection process.

%% file: appendix.tex
\clearpage
\section*{Appendix}

\subsection{Implementation Details}

\textbf{Data Processing.} Following RISE~\cite{rise}, we create the point cloud from a single-view RGB-D image captured by a global camera. Then both the input point clouds and the output actions are aligned in the same camera coordinate. The point cloud is cropped based on the pre-defined robot workspace (notice that the tabletop points remain after cropping). The coordinates are normalized to $[-1,1]$ based on the robot workspace. The gripper width is also normalized to $[-1,1]$ according to the gripper width range.

\textbf{Point Cloud Encoder.} We implement sparse 3D encoder using MinkowskiEngine~\cite{minkowski} with a voxel size of $5\text{mm}$. The sparse 3D encoder outputs a set of $512$-dimensional point feature vectors. The transformer~\cite{transformer} contains $4$ encoding blocks and $1$ decoding block, with $d_\text{model}=512$ and $d_\text{ff}=2048$. The readout token has a dimension of $512$.

\textbf{Force/Torque Encoder.} The high-frequency force/torque observation of the last $T_o = 200$ steps (approximately $2$ seconds given the frequency of $100\text{Hz}$) is encoded via a 3-layer MLP of dimension $(64, 128, 512)$. We use the same transformer architecture as the point cloud encoder to process these force/torque tokens with sinusoidal positioning encoding along the temporal axis. The readout token has a dimension of $512$.

\textbf{Future Contact Predictor.} We utilize a ResNet18~\cite{resnet} vision encoder and a two-layer MLP of dimension $(128, 512)$ to process image and force/torque inputs respectively. The outputs are concatenated and passed through a linear layer to compute the future contact probability $\phi$. The ground-truth future contact state $t$ is generated from the force/torque data within the time window $[t - 2\text{s}, t + 2\text{s}]$ and checks whether force/torque value exceeds the predefined force and torque thresholds. The thresholds may differ across tasks and can be easily determined manually from the collected demonstrations. The ground-truth future contact states are then used to supervise the future contact predictor.

\textbf{Action Head.} A CNN-based diffusion head~\cite{diffusionpolicy} is employed with $100$ denoising iterations for training and $20$ iterations for inference using DDIM~\cite{ddim} scheduler. FoAR predicts the future action of $T_a=20$ steps.

\textbf{Training.} FoAR is trained on 2 NVIIDA A100 GPUs with a batch size of $240$, an initial learning rate of $3\times 10^{-4}$, and a warmup step of $2000$. The learning rate is decayed by a cosine scheduler. During training, we apply the same point cloud augmentations as RISE~\cite{rise}, and we also leverage color jittering for improved robustness. The weighting factor $\alpha$ between the future contact predictor loss and the action loss is set as $0.1$.

\textbf{Deployment.} We apply the reactive control during deployment, combining with simple end-effector position control (\textit{i.e.}, Line 20 in Alg.~\ref{alg:FoAR} is sent to the end-effector position controller of the robot). No advanced control strategies like compliance control, admittance control, and hybrid force/torque control are used in this paper. The future contact probability threshold $\delta_\phi$ is set to $0.9$, the force threshold $\delta_f$ is set to $8\text{N}$, the torque threshold $\delta_t$ is set to $5\text{N}\cdot\text{m}$. The motion direction is calculated based on the predicted future $T_f = 5$ action steps, and we set the small step $\epsilon = 0.006\text{m}$.

\subsection{Additional Ablation}

We conduct an additional ablation experiment by replacing the transformer in the force/torque encoder with a simple MLP, as in the future contact predictor.

\begin{table}[htbp]    
    \centering
    \footnotesize
    \setlength\tabcolsep{2.5pt}
    \begin{tabular}{cccc}    
    \toprule
    \multirow{2}{*}{\textbf{Method}} & \multirow{2}{*}{\textbf{Score} $\uparrow$} & \multicolumn{2}{c}{\textbf{ASR}(\%) $\uparrow$}
    \\ \cmidrule(lr){3-4}
    & & Grasp & Peel \\ 
    \midrule
    RISE~\cite{rise} & 0.293 & \textbf{100} & 50 \\
    FoAR (MLP) & 0.426 & \textbf{100} & 75 \\
    FoAR  & \cellcolor[HTML]{CAD4E7}0.588 & \textbf{100} & \textbf{100} \\
    \bottomrule
    \end{tabular}
    \caption{\textbf{Ablation Results of the \textit{Peeling} Task on Different Force/Torque Encoders.}}
    \label{tab:mlp}
\end{table}

The results shown in Tab.~\ref{tab:mlp} illustrate that the simple MLP encoder cannot effectively capture the temporal features of the force/torque information, resulting in inferior performance compared to the transformer-based encoder. However, it still outperforms RISE by incorporating force/torque information for contact-rich manipulations.